\documentclass[10pt,twocolumn,letterpaper]{article}

\usepackage{iccv}
\usepackage{times}
\usepackage{epsfig}
\usepackage{graphicx}
\usepackage{amsmath}
\usepackage{amssymb}

\usepackage{booktabs}
\usepackage{bbm}

\usepackage[breaklinks=true,letterpaper=true,colorlinks,bookmarks=false]{hyperref}

\iccvfinalcopy 

\def\iccvPaperID{****} 

\ificcvfinal\fi

\begin{document}

\title{Real-time Instance Segmentation with Discriminative Orientation Maps}

\author{\vspace{1.5mm}Wentao Du\quad Zhiyu Xiang\*\thanks{Corresponding author. Email: xiangzy@zju.edu.cn}\quad Shuya Chen\quad Chengyu Qiao\quad Yiman Chen\quad Tingming Bai\\
	College of Information Science \& Electronic Engineering, Zhejiang University
}

\maketitle
\ificcvfinal\thispagestyle{empty}\fi

\begin{abstract}
	Although instance segmentation has made considerable advancement over recent years, it's still a challenge to design high accuracy algorithms with real-time performance. In this paper, we propose a real-time instance segmentation framework termed OrienMask. Upon the one-stage object detector YOLOv3, a mask head is added to predict some discriminative orientation maps, which are explicitly defined as spatial offset vectors for both foreground and background pixels. Thanks to the discrimination ability of orientation maps, masks can be recovered without the need for extra foreground segmentation. All instances that match with the same anchor size share a common orientation map. This special sharing strategy reduces the amortized memory utilization for mask predictions but without loss of mask granularity. Given the surviving box predictions after NMS, instance masks can be concurrently constructed from the corresponding orientation maps with low complexity. Owing to the concise design for mask representation and its effective integration with the anchor-based object detector, our method is qualified under real-time conditions while maintaining competitive accuracy. Experiments on COCO benchmark show that OrienMask achieves 34.8 mask AP at the speed of 42.7 fps evaluated with a single RTX 2080 Ti. The code is available at \url{https://github.com/duwt/OrienMask}.
\end{abstract}

\section{Introduction}

\begin{figure}[t]
	\begin{center}
		\includegraphics[width=0.95\linewidth]{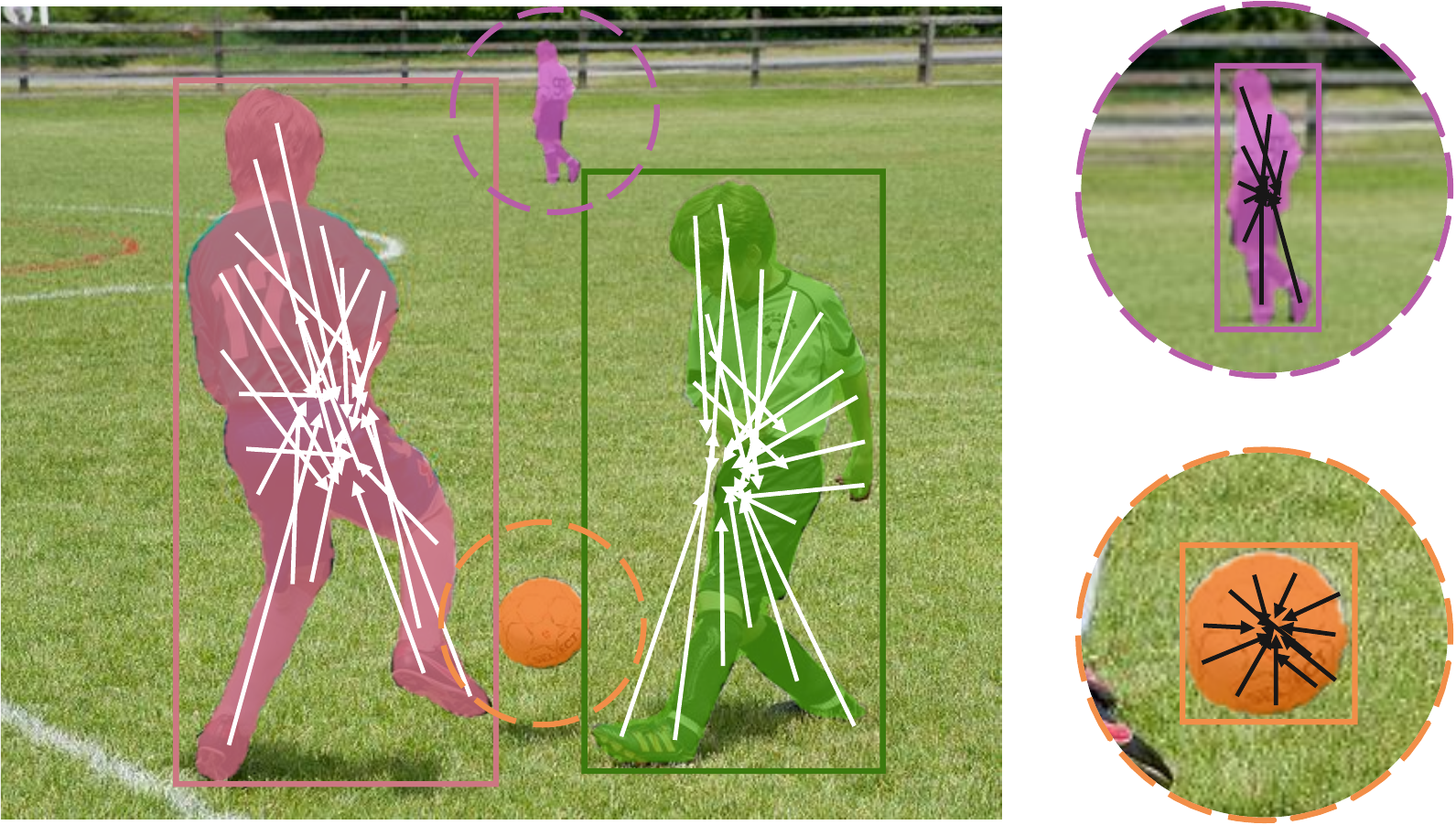}
	\end{center}
	\caption{{\bf Orientation-based Mask Construction.} Those arrowed lines denote densely predicted orientation vectors. Each mask is constructed by gathering all pixels pointing to the central region of the instance in the matched orientation map (white or black).}
	\label{fig:intro}
	\vspace{-0.8em}
\end{figure}

Instance segmentation aims at pixel-wise predictions for every individual object. It integrates instance-level object detection~\cite{ren_faster_2015, redmon_you_2016, liu_ssd_2016, lin_focal_2017} and pixel-level semantic segmentation~\cite{long_fully_2015, chen_deeplab_2017, chen_encoder-decoder_2018}, formulating a more fine-grained visual perception task. Currently there are two dominating types of solutions, namely detection-based and segmentation-based methods. The former extends an object detector with additional foreground dense predictions while the latter deploys specific per-pixel attributes or embeddings to separate instances of the same category in a bottom-up way.

Both paradigms have apparent drawbacks. Conventional detection-based methods like Mask R-CNN~\cite{he_mask_2017} rely on the features pooling operation to project all regions of interest (RoIs) into a fixed size. Since the subsequent mask head should be applied to abundant feature maps of every region proposal, the speed is largely constrained especially when objects densely appear. Moreover, the constant mask resolution brings in unnecessary computation for small objects and loses valuable details for large targets. On the contrary, segmentation-based methods~\cite{liu_affinity_2018, liang_proposal-free_2018} retain the fine-grained appearance and geometry in a pixel-to-pixel manner. They can acquire satisfactory results at elementary scenarios but often fall behind detection-based approaches in accuracy. When the scale of objects varies and the number of categories increases, the generalization of pixel-level clustering adopted in segmentation-based methods is still in doubt.

For the requirement of real-time inference, YOLACT~\cite{bolya_yolact:_2019} is proposed along with a special mask construction scheme, which linearly combines shared non-local prototypes with instance-wise coefficients. It discards the RoI pooling operation that commonly adopted in earlier detection-based methods and directly assemblies masks from fine-grained feature maps. Following this paradigm, an improved approach named BlendMask~\cite{chen_blendmask_2020} is put forward. It replaces the 1D instance-specific coefficients with a set of attention maps, which supply additional spatial-adaptive information to enrich fine-grained details of masks. The success of these solutions shows great potential to incorporate informative global features into detection-based methods. However, one noticeable flaw of these approaches lies in the dependence of RoI cropping operations when generating the assembled masks, which may bring in some mask incompleteness due to inaccurate bounding box predictions.

In this work, we attempt to integrate fine-grained expressions with an one-stage detector in another way. To be specific, we focus on compact mask representation and efficient integration with the anchor-based detector YOLOv3~\cite{redmon_yolov3:_2018} to achieve real-time performance. First, a novel discriminative orientation map is proposed to encode multiple masks independently, where pixels are assigned with centripetal or centrifugal vectors according to their positive or negative labels. This design is totally free from any other semantic segmentation or foreground predictions and is lightweight to decode the complete masks. In addition, considering objects of diverse scales vary the magnitude distributions of orientation vectors, multi-scale design is also taken into account. We assign different orientation maps for instances matching with certain anchor sizes so that the completeness of mask representation is guaranteed. OrienMask merely extends an extra head to the object detector and its function is tightly combined with the box assignment and pre-defined anchor sizes. During inference, for each predicted bounding box, its instance mask can be quickly constructed based on the discriminative vectors in the corresponding orientation map, as shown in Figure \ref{fig:intro}. This process is simple and direct, consisting of nothing but determining binary labels for all pixels by the spatial destinations that orientation vectors indicate. The main contributions of our work can be summarized as follows:
\begin{itemize}
	\item We put forward a light and discriminative orientation-based mask representation for real-time instance segmentation. By defining opposite orientation vectors for foreground and background pixels, we are able to effectively encode multiple instance masks in a fine-grained two-channel map without the need for explicit foreground segmentation. In inference, given the target regions of instances, their masks can be easily constructed from orientation maps in parallel.
	\vspace{-0.5em}
	\item To deal with objects with various sizes, we propose an instance grouping mechanism derived from anchor-based detectors. Each group of instances with similar sizes are assigned to share a common class-agnostic orientation map. We also expand the annotated bounding boxes to provide sufficient supervision for background. The enlarged valid training area not only balances the number of positive and negative samples but also helps distinguishing them around the boarders.
	\vspace{-0.5em}
	\item We integrate the discriminative orientation maps into a fast anchor-based detector YOLOv3 and implement the resulting model OrienMask end-to-end. Experiments show that it is able to achieve 34.8 mask AP at a speed of 42.7 fps on COCO benchmark, which is quite competitive among state-of-the-art real-time methods.
\end{itemize}

\begin{figure*}[t]
	\begin{center}
		\includegraphics[width=0.97\linewidth]{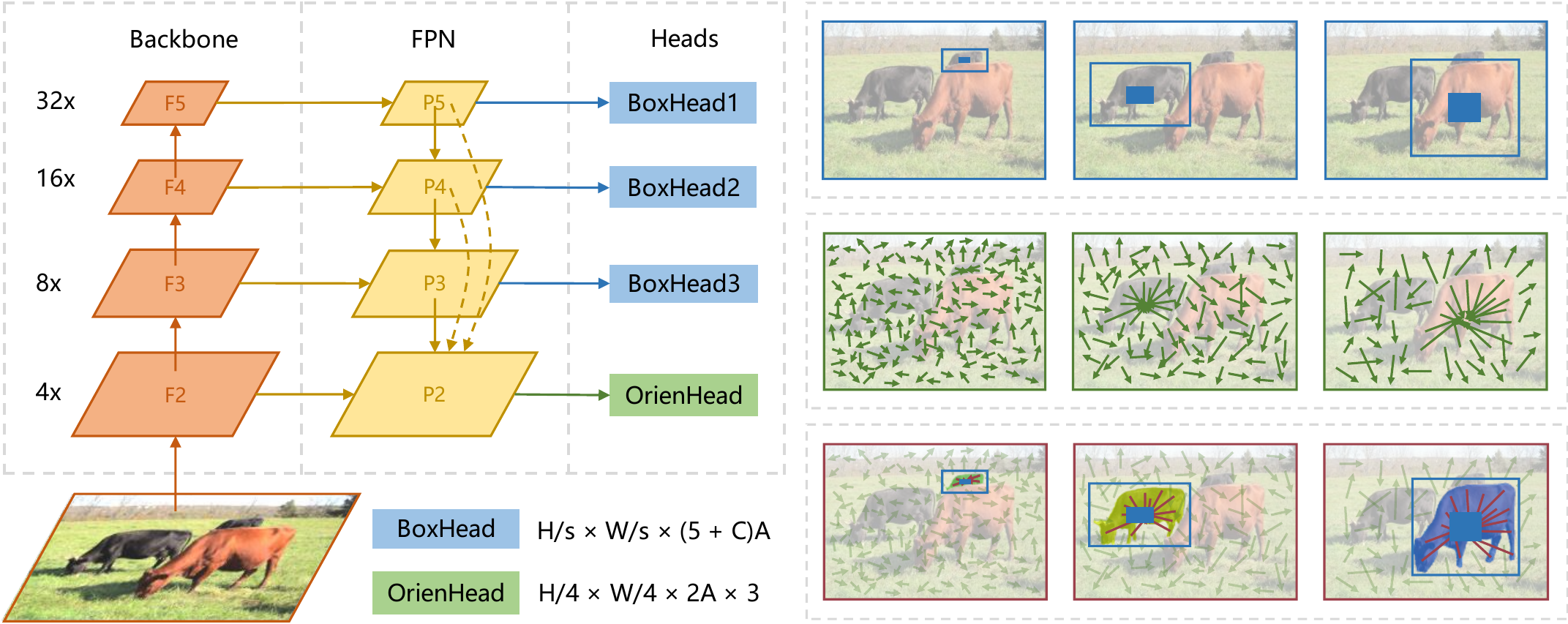}
	\end{center}
	\caption{{\bf OrienMask Architecture.} Left: The network is built upon YOLOv3 with an additional head to predict different orientation maps for each anchor size. $H$ and $W$ are height and width of the input image while $s$ denotes the output stride. There are $C$ categories and $A$ anchors per spatial position. Pyramid features fed to OrienHead are strengthened in our improved version as dashed lines indicate. Right: There are three instances matching with different anchor sizes. The bounding boxes in the first row are predicted by BoxHeads while the orientation maps in the second row are predicted by OrienHead. Each pair in the same column individually determines an instance mask.}
	\label{fig:framework}
	\vspace{-0.8em}
\end{figure*}

\section{Related Work}

\noindent \textbf{Detect-then-segment Methods} With the help of proposals generated by object detectors, detect-then-segment methods first extract reliable RoIs from feature maps, and then obtain fine-grained instance representations. Driven by the success of two-stage detector Faster R-CNN~\cite{ren_faster_2015}, Mask R-CNN~\cite{he_mask_2017} adds a branch for mask prediction in parallel with bounding box regression, and employs RoIAlign to fix the misalignment caused by spatial quantization. After that, PANet~\cite{liu_path_2018} is presented to strengthen message passing in the bottom-up path and fuse pooling features from all levels. HTC~\cite{chen_hybrid_2019} extends Mask R-CNN into a cascade structure by interleaving the mask and box branches while maintaining semantic features fusion. Instead of using the confidence of the detector, Mask Scoring R-CNN~\cite{huang_mask_2019} predicts an extra score to accurately represent the mask quality. However, due to the heavy computations in the second stage, these approaches can hardly satisfy the requirement of real-time inference.

\smallskip
\noindent \textbf{Detect-and-segment Methods} Benefiting from those compact architectures of one-stage object detection~\cite{liu_ssd_2016, lin_focal_2017, redmon_yolov3:_2018, law_cornernet_2018, duan_centernet_2019, tian_fcos_2019}, detect-and-segment methods customize masks from the global feature maps jointly with implicit instance-specific representations. In YOLACT~\cite{bolya_yolact:_2019}, acknowledged as the milestone of this paradigm, a series of mask coefficients are produced along with the box predictions. Then they are multiplied with a set of high-resolution prototypes to generate instance masks. Chen \etal~\cite{chen_blendmask_2020} rethink the trade-off between feature resolution and coefficients dimension, and propose BlendMask, which blends some attention maps of each instance with a group of shared bases. Inspired by conditionally parameterized convolutions~\cite{yang_condconv_2019}, CondInst~\cite{tian_conditional_2020} predicts instance-aware convolutional kernel weights and applies them on high-resolution feature maps. Thanks to the flexible framework and fine-grained representation, these methods keep a good balance between speed and accuracy. Compared with them, our OrienMask employs an explicit and discriminative features sharing scheme for mask representation rather than implicit parameterized forms, which is more concise and provides strong interpretability.

\smallskip
\noindent \textbf{Other Mask Representations} Apart from bounding boxes and foreground probability maps, some compact mask representations also contribute to instance segmentation. For example, Jetley \etal~\cite{jetley_straight_2017} deploy an auto-encoder to compress masks into low-dimensional vectors which can be incorporated in a detector. Xu \etal~\cite{xu_explicit_2019} describe an instance as a series of inner-center radii and encode them into Chebyshev polynomial coefficients. To achieve better precision, polar centerness and polar IoU loss are proposed in PolarMask~\cite{xie_polarmask_2019}. Peng \etal~\cite{peng_deep_2020} implement the snake algorithm in a learning-based form and the circular convolution is proposed to iteratively regress sampling points towards contour positions. Serving as effective representation, pixel offset and its variants are popular in separating instances within the segmented foreground. Uhrig \etal~\cite{rosenhahn_pixel-level_2016} deploy template matching by predicted depth classes and discrete directions to assign pixels of the same semantic label to different instance centers. Box2Pix~\cite{uhrig_box2pix:_2018} matches the foreground pixels with predicted box centers according to offset vectors. Similarly, Li \etal~\cite{li2018pixel} merge foreground pixels based on adaptive voting zones around detection centers. Neven \etal~\cite{neven_instance_2019} propose a joint optimization scheme for class-specific seed and sigma maps along with dense offset vectors. Given the learned clustering bandwidth, masks are sequentially recovered. PersonLab~\cite{papandreou2018personlab} utilizes the short-range and mid-range offset vectors to decode person poses, and then clusters foreground pixels by long-range offsets. Novotny \etal~\cite{ferrari_semi-convolutional_2018} propose a semi-convolutional operator, which adds coordinates to part of learned embedding. PointGroup~\cite{jiang2020pointgroup} extends offset descriptor to 3D instance segmentation, where points of the same category are grouped step by step. Our method also draws inspiration from spatial offset descriptor. However, unlike above methods which mostly use spatial offset to assign segmented foreground pixels to instances, our orientation map is self-discriminative. It is able to filter out background regions and separate instances at the same time. To achieve this goal, special valid training areas are defined and distinctive orientation vectors for both positive and negative samples are considered. Besides, these fine-grained orientation maps are tightly bonded with anchors of the detector, which retains the mask completeness of different sizes and eases the regression.

\section{OrienMask}

\subsection{Overall Architecture}

The network architecture of OrienMask is mainly built upon anchor-based detector YOLOv3~\cite{redmon_yolov3:_2018} with the backbone Darknet-53. As illustrated in Figure \ref{fig:framework}, we add an extra OrienHead to predict orientation maps, which is the key part of the whole framework. The obtained bounding boxes and orientation maps are combined for mask construction.

YOLOv3 deploys 9 bounding box priors and evenly assigns them across 3 scales. Given an image with height $H$ and width $W$ as input, the BoxHead at the scale of output stride $s$ produces $H/s \times W/s \times A$ bounding box predictions, where $A$ denotes the number of anchors per grid cell. OrienHead takes the largest feature maps P2 after feature pyramid network (FPN) as input and then predicts $3A$ orientation maps of fixed resolution $H/4 \times W/4$ with two channels, matching with $3A$ different anchor sizes respectively. Given that handling feature maps of high resolutions is time-consuming, our OrienHead is designed to be light-weight. It is interleaved with three $3\times 3$ and $1\times 1$ convolution layers of input channels 128 and 256 alternatively. After the standard non-maximum suppression, each bounding box is paired with an orientation map according to its anchor size. As demonstrated in the right part of Figure \ref{fig:framework}, all pixels whose orientation vectors end within a contracted bounding box form a foreground mask.

\subsection{Orientation-based Mask Representation}

\begin{figure}[t]
	\begin{center}
		\includegraphics[width=0.92\linewidth]{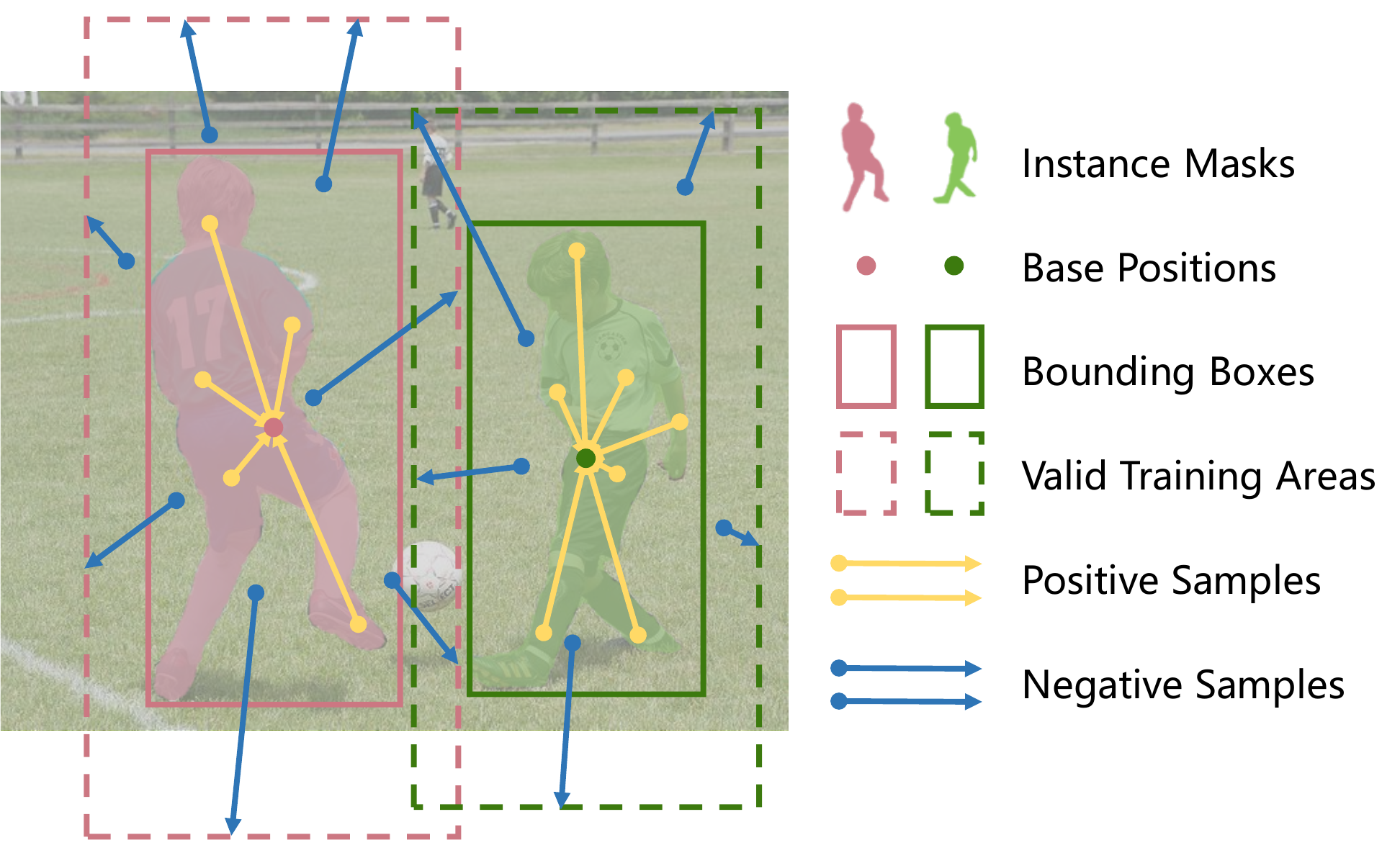}
	\end{center}
	\caption{{\bf Orientation Maps.} For each instance, pixels covered by the mask are positive samples whose orientations are defined as vectors pointing to the base position. Remaining pixels inside the valid training area serve as negative samples, which should point to the border of the valid training area in centrifugal directions.}
	\label{fig:posneg}
	\vspace{-0.3em}
\end{figure}

An orientation map $O^* \in \mathbb{R}^{H\times W\times 2}$ stores pixel-wise spatial offsets in horizontal and vertical directions for multiple instances. Some pivotal concepts at the stage of training are illustrated in Figure \ref{fig:posneg} and will be explained in detail.

\smallskip
\noindent {\bf Valid Training Area} We first expand regions enclosed by annotated bounding boxes to form valid training areas. Pixels outside of any expanded region are ignored during training, which means they are not involved in the loss calculation. All remaining valid pixels are divided into two parts, namely positive and negative samples, based on whether they are covered by instance masks or not. Since the number of positive samples is constant and more negative samples will be counted when the valid training area expands, a proper expand ratio should be determined to balance the number of positive and negative samples. Meanwhile, this expansion also provides sufficient guidance for distinguishing pixels nearby the instance boarders.

\smallskip
\noindent {\bf Orientation Vectors} To easily distinguish positive and negative samples in mask construction process, the orientation vectors of them are defined to point at opposite directions. To be specific, a base position is first specified for each instance and the bounding box centroid is a preferable choice in our experiments. The positive samples on the orientation map are defined pointing to the base position while the negative ones should point to the boarder of the valid training area in centrifugal directions. Denoting the base position as $b$, the target orientation vector $o_i^*$ for pixel $i$ at location $p_i$ can be expressed as
\begin{equation}
o_i^* = \begin{cases}
b - p_i, & \textrm{if positive}, \\
(\alpha_i - 1)(p_i - b), & \textrm{if negative},
\end{cases}
\label{equation:orientation}
\end{equation}
\begin{equation}
\alpha_i = 1 / \max(\dfrac{p_i^x - b^x}{v^l - b^x}, \dfrac{p_i^x - b^x}{v^r - b^x}, \dfrac{p_i^y - b^y}{v^t - b^y}, \dfrac{p_i^y - b^y}{v^b - b^y}).
\end{equation}
The superscripts $x$ and $y$ denote horizontal and vertical directions respectively. Elements in $(v^l, v^r, v^t, v^b)$ represent the leftmost, rightmost, topmost, and bottommost coordinates of the valid training area. All positive samples have the highest priority to override other orientations so that the mask completeness is preserved as much as possible. If a negative sample is overlapped by multiple valid training areas, we simply take the average to avoid ambiguity.

Since orientation vectors are locally defined as spatial offsets pointing to some neighboring destinations, they vary smoothly in both directions within the interior places. The harmony may be slightly disturbed when valid training areas of different instances overlap but the overall unity is still maintained. For each pair of neighboring pixels locating across the mask boundary, the positive is pulled to the base position while the negative is pushed towards outside to the expanded boarder. Thus one direction of gradients at these positions equals to the distance between the base position and the corresponding boarder of valid training area, which is significantly larger than one pixel difference in other interior places. Our experiments in Section \ref{discussion} do prove that the learned orientation maps retain this property, which is helpful to accurately delineate instance masks.

\smallskip
\noindent {\bf Instance Grouping} Although simply stacking all instance masks onto a two-channel orientation map is fascinating for real-time considerations, it could fail in handling some instance overlapping scenarios, such as a person wearing a tie. To alleviate this problem, we introduce an instance grouping mechanism. Noticing that YOLOv3 assigns objects to different anchor sizes based on the intersection over unions with those bounding box priors, we naturally transfer this assignment to our mask representation. To be specific, instances are divided into several groups according to the anchor sizes that they are matched, and each group of instance masks are assigned to an independent orientation map.

Besides solving the overlapping problem caused by objects of different aspect ratios or scales, the instance grouping mechanism has more additional advantages. Since bigger objects often require larger receptive field, this arrangement is beneficial to adapt each orientation map to appropriate scale. Meanwhile, the orientation vectors of grouped instances can be normalized within a small interval so that the magnitude distribution of each group does not vary significantly, which is good for the network training. Moreover, our design also complies with the observation that an image may contain many small instances but only a few large ones. Hence it can preserve as many objects as possible.

\subsection{Mask Construction}

The mask construction process involves two elements: a predicted bounding box $B$ and an orientation map $O$. Recall that each bounding box prediction has an anchor size and each anchor size is associated with an orientation map. Therefore, each bounding box is bound to be matched with an orientation map. Supposing that $B$ and $O$ have been paired, we take the centroid of $B$ as the base position according to the definition in Eq. (\ref{equation:orientation}). Then a rectangular target region centered on the base position is defined, whose size is proportional to the width and height of $B$. If we denote the base position as $b$ and the size of bounding box as $s$, the constructed mask $M$ can be expressed as
\begin{equation}
\begin{split}
M = & \ \left(\left|O_x + P_x - b_x\right| < \tau \cdot s_x\right) \\
& \ \cap \left(\left|O_y + P_y - b_y\right| < \tau \cdot s_y\right).
\end{split}
\end{equation}
Here $P \in \mathbb{R}^{H\times W\times 2}$ stores the coordinates of each pixel, and $\tau$ is a contracting factor to define the target region. In a nutshell, if an orientation vector points to somewhere within certain realms around the base position, the corresponding pixel belongs to the foreground of the given instance. This simple procedure requires no more than point-wise arithmetic or logic operations. For all bounding boxes that survive NMS, the mask construction process can be easily executed in parallel without foreground segmentation. Furthermore, it does not require any RoI cropping operations and directly considers the whole orientation map, resulting little precision leakage caused by inaccurate box predictions.

\subsection{Loss Function}

The loss function consists of two components, providing supervisions for object detection and orientation maps respectively. It can be formulated as
\begin{equation}
\mathcal{L} = \mathcal{L}_{det} + \lambda\mathcal{L}_{orien},
\label{equation:loss}
\end{equation}
where $\lambda$ is a hyper-parameter to balance these two terms. $\mathcal{L}_{det}$ is completely copied from official YOLOv3 without any tricks in the literature.

With regard to $\mathcal{L}_{orien}$, we compute smooth-l1 loss per pixel in valid training areas, and then take the average over positive and negative samples respectively. In addition, we multiply them by the number of instances $N_{inst}$ in accordance with $\mathcal{L}_{det}$. The complete expression is written as 
\begin{equation}
\begin{split}
\mathcal{L}_{orien} = & \ \dfrac{N_{inst}}{N_{pos}}\sum_{i}\mathbbm{1}^\mathcal{P}_i f_{\mathrm{smooth\_l1}}(\dfrac{o_i}{a}, \dfrac{o_i^*}{a}) \\
& \ + \dfrac{N_{inst}}{N_{neg}}\sum_{i}\mathbbm{1}^\mathcal{N}_i f_{\mathrm{smooth\_l1}}(\dfrac{o_i}{a}, \dfrac{o_i^*}{a}),
\end{split}
\end{equation}
where $\mathbbm{1}^\mathcal{P}$ and $\mathbbm{1}^\mathcal{N}$ are the indicator functions for positive and negative samples. $o_i$ denotes the orientation vector predicted at pixel $i$ while $o_i^*$ is the corresponding ground truth. To lift numerical stability and reduce the variance across different scales, $o_i$ and $o_i^*$ are normalized by their anchor size $a$ instead of directly measured in pixels. $\mathcal{L}_{orien}$ for each scale is calculated separately and finally summed up. Though the orientation maps produced by the network is one fourth dimensions of the input image, we calculate and aggregate losses after upsampling them to the full resolution. Likewise, bilinear interpolation is also implemented before feeding orientation maps to construct masks.

\section{Experiments}

We conduct experiments on the challenging MS COCO dataset~\cite{lin_microsoft_2014} and evaluate the predictions with standard metrics. Following the common practice, all models are trained with 118k images of {\tt train2017} and tested on 5k images of {\tt val2017} or 20k images of {\tt test-dev} subset.

\smallskip
\noindent \textbf{Training Details} For the network structure, we retain the official implementation of YOLOv3 and extend a fully convolutional OrienHead as described above. The backbone Darknet-53 is initialized with a pretrained detector and the network is trained end-to-end. We employ stochastic gradient descent (SGD) optimizer with momentum 0.9 and weight decay 0.0005. The batch size is 16 and the synchronized batch normalization is used in our ultimate model but not in ablation study. The initial learning rate is 0.001 and divided by 10 at iterations 520k and 660k respectively. All models are trained for 100 epochs with the input resolution $544 \times 544$. Multiple data augmentations are applied, such as color jitter, random resizing, and horizontal flipping.

\smallskip
\noindent \textbf{Inference Details} Similar to YOLACT~\cite{bolya_yolact:_2019}, input images are directly resized to $544 \times 544$ without test-time augmentation. The inference speed is evaluated on RTX 2080 Ti by default and measured with frames per second (FPS).

\subsection{Ablation Study}

In our ablation experiments, implementations of the object detector is fixed. We adjust other hyper-parameters of our method to obtain the best configurations. To integrate OrienHead with the detector more tightly, some additional refinements will also be applied to the base model.

\smallskip
\noindent \textbf{Valid Training Area} For orientation maps, the definition of negative samples is closely related to the boarder of valid training area. Given that the size of each valid training area is proportional to its bounding box, we vary the expand ratio $r$ from 1.0 to 1.6 with the stride 0.2. As the experimental results shown in Table \ref{table:vta}, our model achieves best performance when $r = 1.2$ and either smaller or larger expand ratio brings some drops in AP. We notice the amount of negative samples and their magnitudes of orientation vectors increase simultaneously as valid training areas expand. A moderate expand ratio keeps a balance between the number of positive and negative samples while maintains enough differentiation around boundaries. It also keeps a proper numeric distribution of negative samples. These two aspects both contribute to better convergence of our model.

\smallskip
\noindent \textbf{Orientation Loss Weight} In the loss function of Eq. (\ref{equation:loss}), $\mathcal{L}_{det}$ and $\mathcal{L}_{orien}$ are used to provide box-level and pixel-level supervisions respectively. In order to associate these two terms together, we explore the orientation loss weight $\lambda$ from 5 to 20 and obtain the results in Table \ref{table:olw}. The mask AP metrics gradually increase as $\lambda$ becomes larger. However, we find the training process gets unstable and the performance saturates when applying weights larger than 20. Hence $\lambda = 20$ is adopted in our subsequent experiments. Moreover, we observe that the adjustment of the orientation loss weight does not disturb the box-level performance remarkably and even has a bit positive effects, which indicates OrienHead maintains the network stability to some extent.

\begin{table}[t]
	\begin{center}
		\begin{tabular}{c@{\hskip 1em}|@{\hskip 1em}*{3}{p{2em}}@{\hskip 1em}|@{\hskip 1em}p{2em}}
			\toprule
			$r$ & AP & AP$_{50}$ & AP$_{75}$ & AP$^{bb}$ \\
			\midrule
			1.0 & 32.0 & 53.4 & 33.1 & 37.0 \\
			1.2 & {\bf 32.5} & {\bf 53.9} & {\bf 33.6} & {\bf 37.4} \\
			1.4 & 32.3 & 53.8 & 33.0 & 37.4 \\
			1.6 & 31.9 & 53.6 & 32.5 & 37.1 \\
			\bottomrule
		\end{tabular}
	\end{center}
	\caption{{\bf Valid Training Area.} An appropriate expand ratio $r$ balances the amounts of positive and negative samples while casting enough supervision to distinguish them.}
	\label{table:vta}
\end{table}

\begin{table}[t]
	\begin{center}
		\begin{tabular}{c@{\hskip 1em}|@{\hskip 1em}*{3}{p{2em}}@{\hskip 1em}|@{\hskip 1em}p{2em}}
			\toprule
			$\lambda$ & AP & AP$_{50}$ & AP$_{75}$ & AP$^{bb}$ \\
			\midrule
			5 & 31.0 & 52.7 & 31.3 & 36.7 \\
			10 & 31.8 & 53.5 & 32.5 & 36.8 \\
			15 & 32.1 & 53.4 & 33.3 & 37.1 \\
			20 & {\bf 32.5} & {\bf 53.9} & {\bf 33.6} & {\bf 37.4} \\
			\bottomrule
		\end{tabular}
	\end{center}
	\caption{{\bf Orientation Loss Weight.} As the orientation loss weight $\lambda$ increases, the mask AP performance of our model is gradually boosted while the box-level AP metric remains relatively stable.}
	\label{table:olw}
	\vspace{-0.5em}
\end{table}

\smallskip
\noindent \textbf{Orientation Target Area} For any bounding box that survives NMS, the mask is simply constructed by collecting all pixels whose orientation vectors point to somewhere close to its base position, without any other operations like RoI cropping. The bounding box is contracted by a scale factor to form an orientation target area so that it is compatible with objects of different aspect ratios or scales. Here we choose contraction ratio $\tau$ from 0.4 to 0.8. From Table \ref{table:ota} we find the performance is sensitive to the contraction ratio and the best performance is obtained when $\tau = 0.6$. More specific parameter tuning for each group of instances is optional to achieve even higher AP.
\begin{table}[t]
	\begin{center}
		\begin{tabular}{c@{\hskip 1em}|@{\hskip 1em}*{3}{p{2em}}}
			\toprule
			$\tau$ & AP & AP$_{50}$ & AP$_{75}$ \\
			\midrule
			0.4 & 30.4 & 50.7 & 31.3 \\
			0.5 & 32.2 & 52.9 & 33.2 \\
			0.6 & {\bf 32.5} & {\bf 53.9} & {\bf 33.6} \\
			0.7 & 31.7 & 53.8 & 32.4 \\
			0.8 & 30.1 & 52.8 & 29.9 \\
			\bottomrule
		\end{tabular}
	\end{center}
	\caption{{\bf Orientation Target Area.} The contraction ratio $\tau$ for orientation target area has strong effects on overall performance.}
	\label{table:ota}
\end{table}
\begin{table}[t]
	\begin{center}
		\begin{tabular}{l@{\hskip 1em}|@{\hskip 1em}*{3}{p{2em}}}
			\toprule
			Method & AP & AP$_{50}$ & AP$_{75}$ \\
			\midrule
			base & 32.5 & 53.9 & 33.6 \\
			+ box centroid & 33.3 & 54.8 & 34.4 \\
			+ larger anchors & 33.8 & 55.2 & 35.2 \\
			+ fpn plus & {\bf 34.1} & {\bf 55.7} & {\bf 35.4} \\
			\bottomrule
		\end{tabular}
	\end{center}
	\caption{{\bf Other Improvements.} `+ box centroid' means replacing grid centers with box centroids as base positions for orientation maps. `+ larger anchors' inherits anchor settings from YOLOv4 which have about 1.2 times the sizes of the original version. `+ fpn plus' merges pyramid features of P3, P4, and P5 to generate P2 for better multi-scale expressions in OrienHead.}
	\label{table:om}
	\vspace{-0.6em}
\end{table}

\begin{table*}[t]
	\begin{center}
		\begin{tabular}{l@{\hskip 2em}ll@{\hskip 2em}l@{\hskip 2em}*{6}{p{2em}}}
			\toprule
			Method & Backbone & Size & FPS & AP & AP$_{50}$ & AP$_{75}$ & AP$_{S}$ & AP$_{M}$ & AP$_{L}$ \\
			\midrule
			Mask R-CNN*~\cite{he_mask_2017} & ResNet-50 & 800$\times$ & 18.5 & 37.5 & 59.3 & 40.2 & 21.1 & 39.6 & 48.3 \\
			BlendMask~\cite{chen_blendmask_2020} & ResNet-50 & 800$\times$ & 17.6 & 37.0 & 58.9 & 39.7 & 17.3 & 39.4 & 52.5 \\
			CondInst~\cite{tian_conditional_2020} & ResNet-50 & 800$\times$ & 17.9 & 37.8 & 59.1 & 40.5 & 21.0 & 40.3 & 48.7 \\
			SOLO~\cite{wang_solo:_2019} & ResNet-50 & 800$\times$ & 11.2 & 36.8 & 58.6 & 39.0 & 15.9 & 39.5 & 52.1 \\
			PolarMask~\cite{xie_polarmask_2019} & ResNet-101 & 800$\times$ & 12.3* & 32.1 & 53.7 & 33.1 & 14.7 & 33.8 & 45.3 \\
			MEInst~\cite{zhang_mask_2020} & ResNet-101 & 800$\times$ & 12.8* & 33.9 & 56.2 & 35.4 & 19.8 & 36.1 & 42.3 \\
			\midrule
			YOLACT~\cite{bolya_yolact:_2019} & ResNet-101 & 550 & 38.5 & 29.8 & 48.5 & 31.2 & 9.9 & 31.3 & 47.7 \\
			CenterMask~\cite{wang_centermask_2020} & DLA-34 & 512 & 25.2* & 33.1 & 53.8 & 34.9 & 13.4 & 35.7 & 48.8 \\
			{\bf OrienMask} & Darknet-53 & 544 & 42.7 & 34.8 & 56.7 & 36.4 & 16.0 & 38.2 & 47.8 \\
			\bottomrule
		\end{tabular}
	\end{center}
	\caption{{\bf Quantitative results} on COCO {\tt test-dev}. We compare OrienMask with some typical frameworks. The speed is evaluated on 1k samples with the same platform except for those marked with `*' whose statistics are inferred from their publications. Mask R-CNN model comes from {\tt Detectron2}. The input size is square by default and the notation ended with `$\times$' means the length of shorter side.}
	\label{table:qr}
\end{table*}

\begin{table*}[t]
	\begin{center}
		\begin{tabular}{l@{\hskip 2em}llc@{\hskip 2em}c@{\hskip 2em}*{6}{p{2em}}}
			\toprule
			Method & Backbone & Size & Space & FPS & AP & AP$_{50}$ & AP$_{75}$ & AP$_{S}$ & AP$_{M}$ & AP$_{L}$ \\
			\midrule
			YOLACT~\cite{bolya_yolact:_2019} & Darknet-53 & 550 & 4.52M & 45.9 & 28.9 & 46.9 & 30.3 & 9.8 & 30.9 & 47.3 \\
			YOLACT++~\cite{bolya_yolact_2019} & ResNet-50 & 550 & 9.04M & 40.8 & 33.7 & 52.7 & 35.5 & 11.9 & 36.6 & 54.6 \\
			BlendMask~\cite{chen_blendmask_2020} & ResNet-50 & 550$\times$ & 18.44M & 35.6 & 34.5 & 54.7 & 36.5 & 14.4 & 37.7 & 52.1 \\
			MEInst~\cite{zhang_mask_2020} & ResNet-50 & 512 & 1.41M & 28.1 & 32.2 & 53.9 & 33.0 & 13.9 & 34.4 & 48.7 \\
			SOLO-Lite~\cite{wang_solo:_2019} & ResNet-50 & 512$\times$ & 2.95M & 29.7 & 33.0 & 53.9 & 34.9 & 11.5 & 37.0 & 51.5 \\
			SOLOv2-Lite~\cite{wang_solov2_2020} & ResNet-50 & 448$\times$ & 7.25M & 39.6 & 33.7 & 53.3 & 35.6 & 11.3 & 36.9 & 55.4 \\
			{\bf OrienMask} & Darknet-53 & 544 & 1.27M & 41.9 & 34.5 & 56.0 & 35.8 & 16.8 & 38.5 & 49.1 \\
			\bottomrule
		\end{tabular}
	\end{center}
	\caption{{\bf Real-time instance segmentation} on COCO {\tt val2017}. With source codes and trained weights released by their authors, all methods are evaluated in our platform. `Space' means the memory occupation of feature maps in megabytes that used to construct masks.}
	\label{table:rtis}
	\vspace{-0.6em}
\end{table*}

\smallskip
\noindent \textbf{Other Improvements} We further explore some measures to better integrate OrienHead with the detector and improve the overall performance. These measures are applied step by step and the results are illustrated in Table \ref{table:om}. For the orientation definition, we initially choose the grid centers as base positions, which keeps consistent with the rule of box regression in the detector. Since the box centroid is more accurate to locate an instance, we adopt it as the base position. This refinement improves the mask AP metric from 32.5 to 33.3. Then we are inspired by the next generation of YOLO framework~\cite{bochkovskiy_yolov4_2020} to adopt larger anchors, which are proved more suitable for the given input resolution. It should be noted that no extra training tricks are taken and we still maintain other settings as pure YOLOv3. Benefiting from the larger anchors, the performance is boosted by 0.5 AP. In earlier experiments, we follow the standard FPN to produce P2 for OrienHead but the predicted orientation maps are related with box predictions from multiple scales. To closely associate these two outputs, we merge multi-scale pyramid features to predict orientation maps while keeping the same network structure in subsequent layers, as dashed lines in Figure \ref{fig:framework} indicate. The resulting model surpasses the previous once again with negligible extra computing cost.

\begin{figure*}[t]
	\begin{center}
		\includegraphics[width=0.8\linewidth]{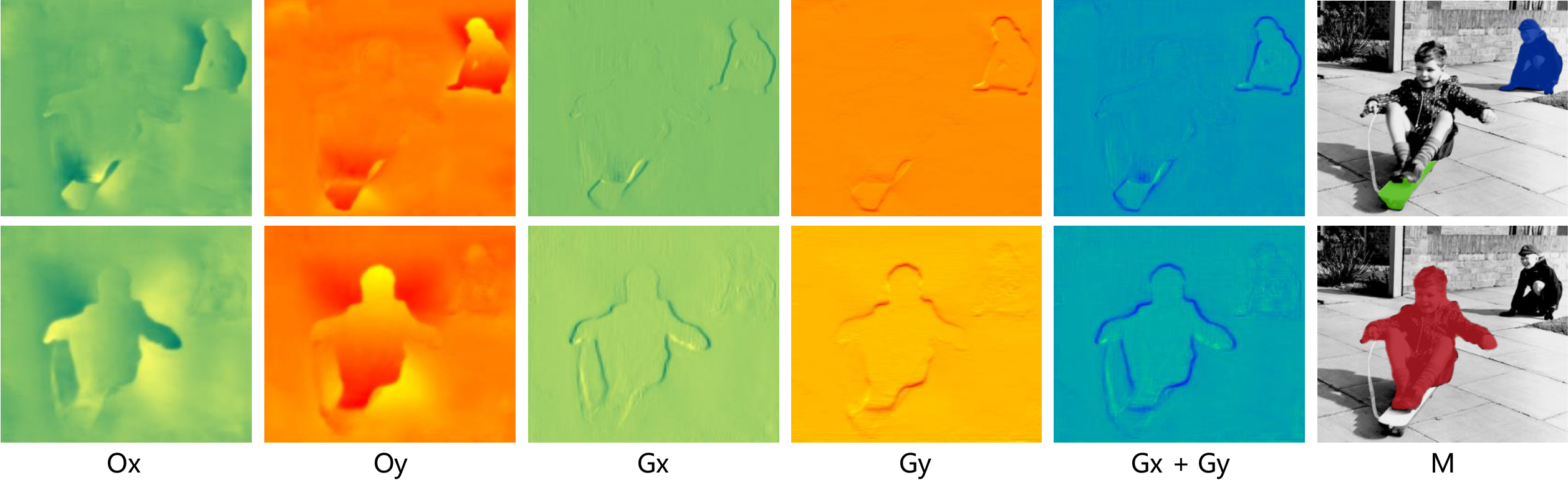}
	\end{center}
	\caption{{\bf Orientation maps (O) and their gradients (G)} for different anchor sizes at two directions (x and y), where the brighter colors mean the larger signed values. Masks (M) constructed by each orientation map are filled with distinct colors and displayed separately.}
	\label{fig:orien}
\end{figure*}

\begin{figure*}[t]
	\begin{center}
		\includegraphics[width=0.905\linewidth]{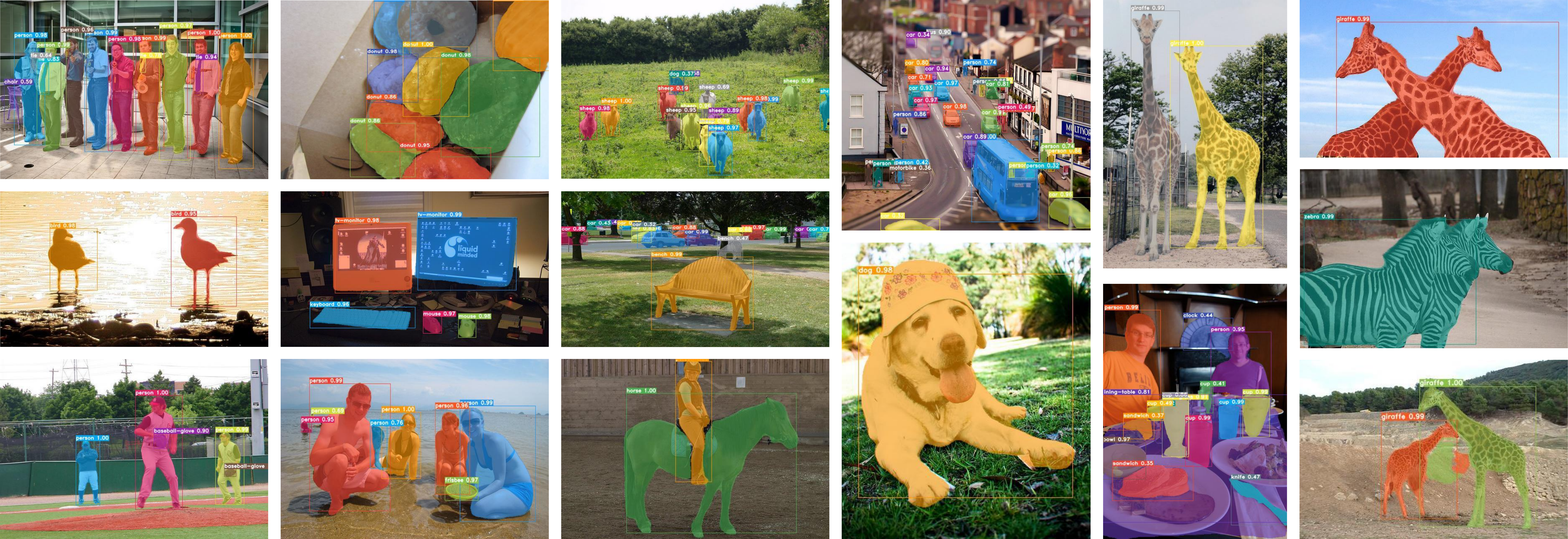}
	\end{center}
	\caption{{\bf Qualitative results} of OrienMask on COCO {\tt test-dev}. Predictions with confidence greater than 0.3 are displayed. Our model is able to handle most complex scenarios with satisfactory precision. Two typical failure cases are also shown in the last column.}
	\label{fig:visualization}
	\vspace{-0.6em}
\end{figure*}

\subsection{Comparison with State-of-the-art Methods}

We first evaluate our OrienMask on the canonical COCO {\tt test-dev} benchmark and select a series of representative frameworks for comparison. From the quantitative results displayed in Table \ref{table:qr}, we find OrienMask is both faster and more accurate when compared against those having similar input resolutions like YOLACT and CenterMask, and some methods that aim at simplified mask representation like PolarMask and MEInst. We admit that OrienMask falls behind some non-real-time approaches with either higher input image resolutions or more complicated pipelines. When considering the twice or even three times faster in inference, it seems reasonable to accept some sacrifice in accuracy.

Since real-time instance segmentation is the main motivation of our work, we further compare OrienMask with state-of-the-art methods capable of real-time inference on COCO {\tt val2017}. All models adopt relatively shallow backbones along with small input resolutions, which makes the comparison fair and persuasive. As shown in Table \ref{table:rtis}, Our method surpasses YOLACT with 5.6 AP at the cost of 4.0 fps slower. Except for that, OrienMask serves as the leading method in speed comparison and outperforms most counterparts in the mask AP metric. It reaches a good balance between efficiency and accuracy. We also calculate the memory occupation of top feature maps that used for mask construction, and record the results in the `space' column. The input resolution of all methods is assumed to be fixed as $544 \times 544$ for brevity. The statistics manifest that our method occupies the least memory resources to construct masks, which proves its success in reducing redundancy while maintaining good mask quality.

\subsection{Discussions} \label{discussion}

In this subsection, we analyze some underlying properties of our method from a qualitative view. The advantages and limitations are both covered.

\smallskip
\noindent \textbf{Orientation Maps} We pick two predicted orientation maps to unearth the mechanism of our mask representation. As shown in Figure \ref{fig:orien}, the attention of each orientation map is grabbed by objects with the specific anchor size. Two kids and a skateboard are assigned to two orientation maps based on their sizes instead of categories. We display the gradient maps of both directions and their pixel-wise sum, which exhibit the obvious difference around instance boundaries. It can also be observed that two components of gradient maps concentrate on different regions of instances, \ie, the left and right parts are highlighted in $Gx$ while the top and bottom parts are accentuated in $Gy$. Combining these two complementary orientation maps together, the complete contours of objects can be depicted and then all inner pixels for each instance are safely gathered. These visualized patterns along with high-quality predicted masks verify the effectiveness of our orientation-based mask representation.

\smallskip
\noindent \textbf{Qualitative Results} As displayed in Figure \ref{fig:visualization}, our method performs well in separating adjacent instances and precisely delineating their masks. Getting rid of RoI cropping and directly collecting foreground pixels according to the vectors in orientation maps, our mask construction procedure has much tolerance to inaccurate bounding box predictions. Meanwhile, our method also performs well in some complex object overlapping scenarios, especially when one or more small objects locates on a large one. This is illustrated in several images of Figure \ref{fig:visualization}, such as persons wearing ties, food placed on the table, baseball gloves at the front of people and so on. Thanks to the instance grouping mechanism, objects matched with different anchor sizes do not disturb each other and their masks are completely preserved. Moreover, we do not introduce any pixel-level categorical information for OrienHead. The class-agnostic orientation maps are proved to be qualified for recovering masks with satisfactory quality, no matter what categories they belong to.

\smallskip
\noindent \textbf{Failure Cases} Although OrienMask works well for most cases, we observe two typical failures as shown in the last column of Figure \ref{fig:visualization}. The first case appears when two instances with the same category and similar size heavily overlap. Orientation maps cannot distinguish them because pixels of both masks point to almost the same base position. The second failure happens due to the severe confrontation of some background pixels between instances, especially when the base positions locate close to their mask boundaries. For example, two giraffes in the bottom right corner of Figure \ref{fig:visualization} both tend to push the intermediate part outwards, which accidentally makes some background pixels intrude into another target region by mistake. Overall, these errors caused by the incompleteness of mask representation are unusual and only occur in limited cases.

\section{Conclusion}

In this work, a real-time instance segmentation framework termed OrienMask is proposed, which integrates discriminative orientation maps with an anchor-based detector. Apart from those centripetal vectors for foreground pixels, we further consider negative samples in orientation maps so that both background filtering and instances separation can be accomplished at the same time. An instance grouping mechanism is also presented and each orientation map specializes in grouped objects with the same anchor size. Given the target regions indicated by predicted boxes, masks can be efficiently constructed from corresponding orientation maps, without the need for explicit foreground predictions. Experiments on COCO show that the proposed OrienMask can reach competitive accuracy under real-time conditions.

\subsection*{Acknowledgment}
This work is supported by  National Natural Science Foundation of China-Zhejiang Joint Fund
for the Integration of Industrialization and Informatization
(U1709214), and Key Research \& Development Plan of Zhejiang Province (2021C01196).

{\small
	\bibliographystyle{ieee_fullname}
	\bibliography{orienmask}
}

\end{document}